\documentclass[conference]{IEEEtran}
\IEEEoverridecommandlockouts
\usepackage{cite}
\usepackage{stfloats}
\usepackage{amsmath,amssymb,amsfonts}
\usepackage{algorithmic}
\usepackage{graphicx}
\usepackage{textcomp}
\usepackage{xcolor}
\usepackage{color}
\usepackage{hyperref}
\usepackage{orcidlink} 
\usepackage{booktabs}
\usepackage{multirow}
\usepackage{soul}
 \usepackage[normalem]{ulem}

\def\BibTeX{{\rm B\kern-.05em{\sc i\kern-.025em b}\kern-.08em
    T\kern-.1667em\lower.7ex\hbox{E}\kern-.125emX}}

\title{Classifying Healthy and Defective Fruits with a Multi-Input Architecture and CNN Models\\
}

\author{
\IEEEauthorblockN{Luis Chuquimarca}
    \IEEEauthorblockA{\textit{CIDIS} \\
    \textit{ESPOL Polytechnic University}\\
    Guayaquil, Ecuador \\
    lchuquim@espol.edu.ec,\\
    and\\
    \textit{FACSISTEL} \\
    \textit{UPSE Santa Elena Peninsula State University}\\
    La Libertad, Ecuador}
\and
\IEEEauthorblockN{Boris Vintimilla}
\IEEEauthorblockA{\textit{CIDIS} \\
\textit{ESPOL Polytechnic University}\\
Guayaquil, Ecuador \\
boris.vintimilla@espol.edu.ec}
\and
\IEEEauthorblockN{Sergio Velastin}
\IEEEauthorblockA{\textit{School of EECS} \\
\textit{Queen Mary University of London} \\
London, UK\\
sergio.velastin@ieee.org,\\
and\\
\textit{Dept. of Computer Science}\\
\textit{University Carlos III}\\
Madrid, Spain}
}

\begin{document}
\maketitle
\makeatletter
\def\footnoterule{\kern-3\p@
  \hrule \@width 2in \kern 2.6\p@} 
\makeatother
\newcommand{\copyrightnotice}[1]{{%
  \renewcommand{\thefootnote}{}
  \footnotetext[0]{#1}%
}}
\copyrightnotice{979-8-3503-7565-7/24/\$31.00 ©2024 IEEE \newline © 2024 IEEE. Personal use of this material is permitted. Permission from IEEE must be obtained for all other uses, in any current or future media, including reprinting/republishing this material for advertising or promotional purposes, creating new collective works, for resale or redistribution to servers or lists, or reuse of any copyrighted component of this work in other works.  

DOI: https://doi.org/10.1109/10677833
}

\begin{abstract}
This study presents an investigation into the utilization of a Multi-Input architecture for the classification of fruits (apples and mangoes) into healthy and defective states, employing both RGB and silhouette images. The primary aim is to enhance the accuracy of CNN models. The methodology encompasses image acquisition, preprocessing of datasets, training, and evaluation of two CNN models: MobileNetV2 and VGG16. Results reveal that the inclusion of silhouette images alongside the Multi-Input architecture yields models with superior performance compared to using only RGB images for fruit classification, whether healthy or defective. Specifically, optimal results were achieved using the MobileNetV2 model, achieving 100\% accuracy. This finding suggests the efficacy of this combined methodology in improving the precise classification of healthy or defective fruits, which could have significant implications for applications related to external quality inspection of fruits.
\end{abstract}


\section{Introduction}
In today's post-harvest processes, fruit classification is carried out through quality inspection conducted by specialized agents, who consider parameters such as ripeness, deforms, and defects~\cite{palumbo2023computer,liu2023application}. The considerations adhere to international standards for defects, as attributed by regulatory bodies governing international norms. This process is traditionally performed manually, and due to the commercial value attributed to fruits based on their quality level, higher quality implies higher profits and access to better markets~\cite{lu2023survey}. Additionally, both small-scale retailers and supermarkets, as well as consumers, take various factors into account in the purchase and sale of these foods. However, visual appeal is a paramount factor in fruit selection. In the current global market, ensuring the quality control of agricultural products is paramount, with indispensable international standards for producing high-quality products essential for human health~\cite{chuquimarca2023banana,devika2023real}. For this study, the considered parameter is fruit defects (using apples and mangoes), which aids us in classifying whether a fruit is healthy or defective.

In recent years, there has been a significant advancement in the development of automated systems based on computer vision technology to replace manufacturing operations~\cite{chakraborty2023development,dhanush2023comprehensive}. Specifically, the classification of healthy or defective fruits has been addressed using convolutional neural network (CNN) models, which utilize fruit images as input~\cite{li2023smart,nguyen2021intelligent}. These images are evaluated using a series of weights derived from the training characteristics of the CNN model. The primary types of images in this context are binary and true color images, representing the predominant forms of digital images. To obtain specific information from a digital image, an essential process called segmentation is employed~\cite{naranjo2020review,gururaj2023deep}. Following this, a name or label is assigned to the image based on certain visual criteria, with the two classes of healthy or defective fruits being considered in this study~\cite{ibrahim2022cnn,buyukarikan2023classification}. In this research, two simple CNN models, MobileNetV2 and VGG16, are used and analyzed.

The key contributions of this scientific article are as follows:

\begin{itemize}

\item Improvement of the performance of CNN models used in the study~\cite{pacheco2023fruit} by exclusively utilizing real images, no longer involving the generation of synthetic images.

\item Implementation of image preprocessing through fruit segmentation in silhouette format, which is a well-known shape descriptor. This technique allows for a more detailed observation of the presence of defects (rot, bruises, scabs, and black spots) in fruits, significantly enhancing the accuracy of the CNN classification model.

\item Introduction of a Multi-Input architecture that leverages both the features of RGB (Red-Green-Blue) images and silhouettes respectively in each branch. This innovative technique surpasses the results obtained in a previous study~\cite{pacheco2023fruit}, which relied solely on a Single-Input architecture. The utilization of the Multi-Input architecture enables a better capture of the distinctive features of fruits, resulting in higher classification accuracy.

\item The methodology employed in this study integrates the utilization of a dataset comprised of RGB images alongside their corresponding silhouettes to feed into a Multi-input architecture. This architecture undergoes training utilizing both the MobileNetV2 and VGG16 models. As a result, a remarkable precision of 100\% was attained in the classification of both healthy and defective apples using the MobileNetV2 model. Similarly, in the classification of healthy and defective mangoes, the MobileNetV2 model achieved an impressive precision of 99.53\%. These results stand in favorable comparison to those reported in the previous study~\cite{pacheco2023fruit}, where the MobileNetV2 model yielded a precision of 97.50\% for apples and 92.90\% for mangoes. This comparative analysis illuminates the relative efficacy of various architectures in this domain, offering invaluable insights for future research endeavors.

\end{itemize}

The structure of this article is as follows: Section \ref{sec:related works} provides a comprehensive review of the literature related to the classification of healthy and defective fruits. Subsequently, Section \ref{sec:proposed approach} details the proposed methodology for conducting the study. Section \ref{sec:experimental results} showcases the results obtained by the CNN (Multi-Input) models. Finally, in Section \ref{sec:conclusions}, the conclusions derived from this work are presented.

\section{Related Works}
\label{sec:related works}

This section outlines previous studies related to the use of CNN models for classifying healthy or defective fruits, providing an overview of the criteria to consider for the current research. 
Common defects that may manifest in fruits include bruising, spots, and rot. These anomalies, often attributable to inadequate storage conditions or damage during handling, have a significant impact on the sensory and nutritional quality of fruits. The accurate classification of defective fruits is essential for the food industry, as it allows for the separation of products of optimal quality from those that may be deemed unfit for human consumption. In this context, the application of CNN models for the automatic classification of healthy or defective fruits has emerged as a promising solution, offering an efficient and objective assessment of fruit quality at various stages of the supply chain.

The study conducted by~\cite{buyukarikan2023classification} employs deep learning techniques to identify physiological disorders (bitter pit, shriveling, superficial scald) in apples. Leveraging CNN models, which enable efficient feature extraction from fruit images, they classify images of apples affected by post-harvest physiological disorders. Additionally, they highlight the Xception model with the best performance, achieving an accuracy, precision, recall, and F1-score average of 0.996, 0.994, 0.996, and 0.998 respectively, outperforming other models such as ResNet and MobileNet. However, it is important to note that the study does not address the detection of additional defects (bruising, spots, rot) in apples, nor does it utilize color balance models to enhance images, or explore the use of architectures or techniques to strengthen CNN models.

The study reported in~\cite{agarwal2022differential} focuses on detecting diseases in apples, such as blotch, scab, and rot, using deep learning and machine learning approaches. A CNN model with three convolution layers was utilized to classify healthy and defective apples. Additionally, pre-trained models were compressed using Differential Evolution, achieving a maximum compression rate of 82.19\% for the VGG16 model without significant performance loss, thus facilitating its implementation on mobile devices. Binary classification of fruits with defects, alongside a healthy class, proves suitable for deep learning techniques. The proposed CNN model achieved a classification accuracy of 93.75\%, while the best performance among machine learning classifiers was 99\% using Random Forest, although its functionality is specifically geared towards detection. However, it is important to note that the research does not address disease detection in other fruits.

The aim of the research in~\cite{ibrahim2022cnn} is to construct and evaluate various CNN models, including GoogLeNet, AlexNet, and a proposed CNN model, to categorize fresh and rotten appearances using a dataset of apple images. The results obtained reveal that GoogLeNet achieved an accuracy of 100\%, surpassing AlexNet, which achieved 99.69\%, and the proposed CNN model, which reached 87.24\%. However, it is worth noting that a more thorough review of the dataset and an increase in the number of images (greater than 1000 images per class) could allow for the generalization of findings and the development of even more robust CNN models.

The study~\cite{pacheco2023fruit} aims to evaluate different CNN models in identifying various defects in apples and mangoes to ensure the quality of fruit production. Utilizing the InceptionV3, MobileNetV2, VGG16, and DenseNet121 models, trained with a dataset comprising real and synthetic images of apples and mangoes, as well as both healthy and defective fruits exhibiting rot, bruises, scabs, and black spots. MobileNetV2 demonstrated the highest accuracy, achieving 97.50\% for apples and 92.50\% for mangoes, making it the optimal choice for classifying healthy or defective fruits. In contrast, InceptionV3 and DenseNet121 exhibited accuracy above 90\%, while VGG16 had a less satisfactory performance, failing to surpass 80\% accuracy for either type of fruit. The robust dataset comprises 20k images (10k for each fruit), encompassing real and synthetic images. However, there is a need to enhance the CNN classifier model for defect classification and explore advanced architectures as potential future research directions.

Another article~\cite{pachon2020fruit} introduces a proprietary algorithm based on deep learning designed for the identification and assessment of the state of fruits, distinguishing between those that are healthy and those with defects. Eight types of fruits were addressed, resulting in a total of 16 categories for classification. A CNN model with a Directed Acyclic Graph (DAG) structure was proposed for fruit recognition and classification. This CNN model demonstrated an accuracy of 94.43\% in classifying 1600 test images, with processing times of approximately 45 to 55 milliseconds. The DAG structure of the CNN allows each branch to learn characteristics of different sizes, enabling increased network depth without compromising processing speed. However, it lacks an analysis of the types of defects present in the fruits, as well as the standards considered for conducting quality inspection thereof.

In ~\cite{nithya2022computer}, a computer-assisted classification system has been presented for detecting defects in mangoes using a deep learning approach. Specifically, the CNN model is employed with a publicly accessible mango dataset as input. Experimental results reveal that the proposed method achieved an accuracy of 98\%, tested on a set of 800 mango images. The proposed model has not been compared with other publicly available models, which could be an aspect to consider in future research.

Following a thorough review of the state of the art, four types of defects commonly found in apples and mangoes were identified, which are categorized into the following groups: rot, bruises, scabs, and black spots~\cite{pacheco2023fruit}. In the specific case of apples, studies such as that conducted by~\cite{buyukarikan2023classification} identify three types of defects such as bitter pit classified within the black spots category, shriveling associated with rot, and superficial scald belonging to the scabs category. Conversely, research such as that carried out by~\cite{agarwal2022differential} identifies three types of defects a blotch classified within the scabs category, scab, and rot.

Upon reviewing studies~\cite{pachon2020fruit} and ~\cite{nithya2022computer}, which investigate defects in mangoes, it is evident that the specific types of defects are not detailed. In contrast, study~\cite{pacheco2023fruit} explicitly identifies defects such as rot, bruises, scabs, and black spots, which are also addressed in the present work.

After examining the reviewed studies, we observe a trend toward strengthening existing CNN models by exploring available datasets and implementing novel architectures to enhance their performance. This leads to obtaining superior results in the evaluation metrics of the CNN models.

\section{Proposed Approach}
\label{sec:proposed approach}

This section presents the dataset acquisition along with its respective preprocessing, which includes generating silhouette images using segmentation techniques. The proposed approach aims to enhance the accuracy of CNN models in binary classification of apples, distinguishing between healthy and defective ones, which include bruises, blemishes, rot, and black spots. Therefore, we propose two strategies in this work. The first one is based on using silhouette images to complement RGB images, resulting in improved binarization of the dataset. The second strategy takes advantage of the first by incorporating it into the Multi-Input architecture where in one of its branches silhouette images are used, while in the other the corresponding RGB images are used, both serving as inputs for the CNN models used. The following sections provide more details of the proposed methodology.

\subsection{Data Acquisition}

The input dataset was sourced from a public platform (\url{https://github.com/luischuquim/Healthy-Defective-Fruits}), encompassing a combination of real and synthetic images. However, only 10k real images of apples and mangoes were utilized. Notably, the dataset is carefully labeled, ensuring an equitable distribution between healthy and defective fruits for each fruit type.

\begin{figure}[htbp]
\centerline{\includegraphics[width=0.7\columnwidth]{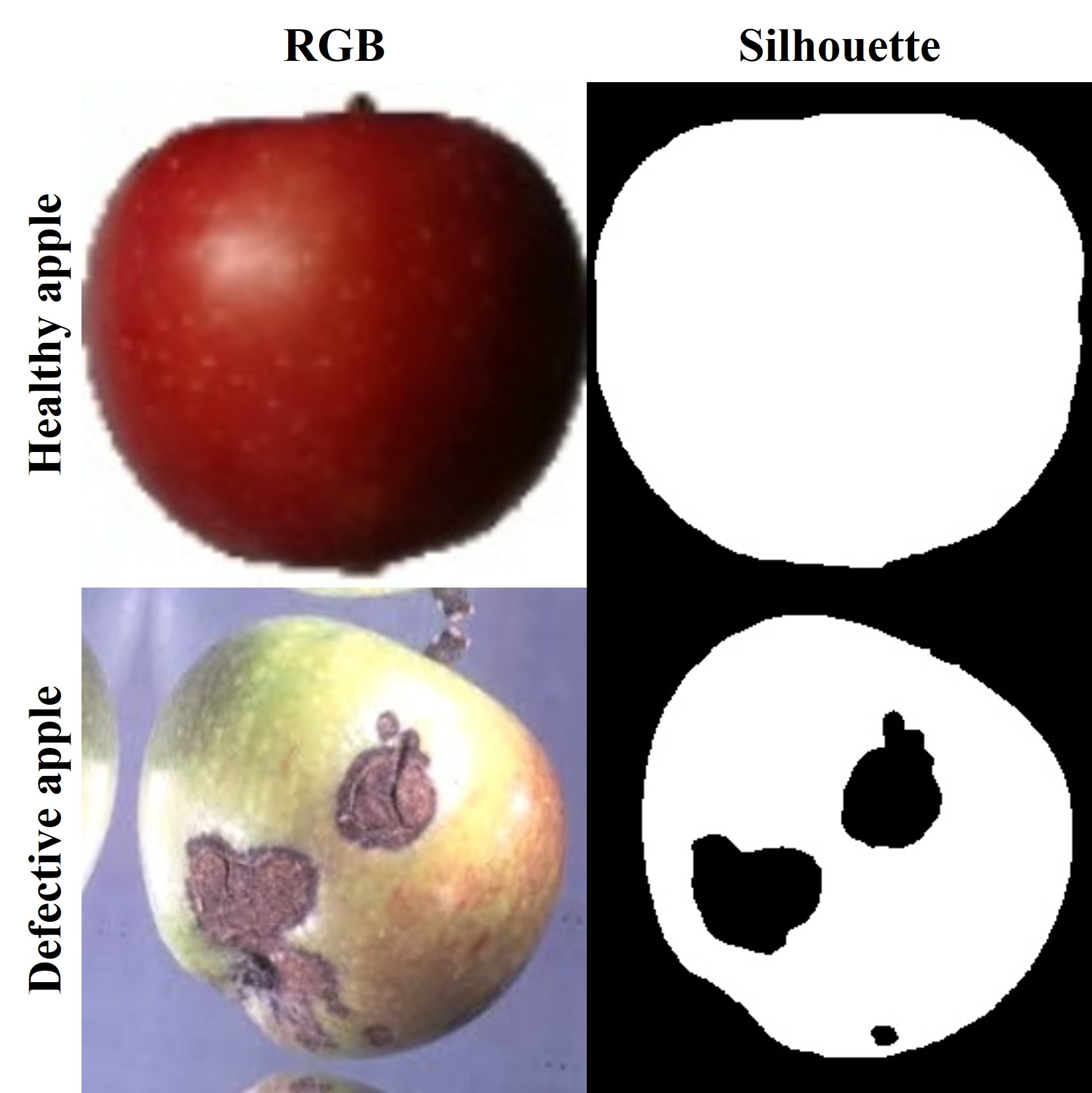}}
\caption{Apple silhouette.}
\label{fig_apple_sil}
\end{figure}

\begin{figure}[htbp]
\centerline{\includegraphics[width=0.7\columnwidth]{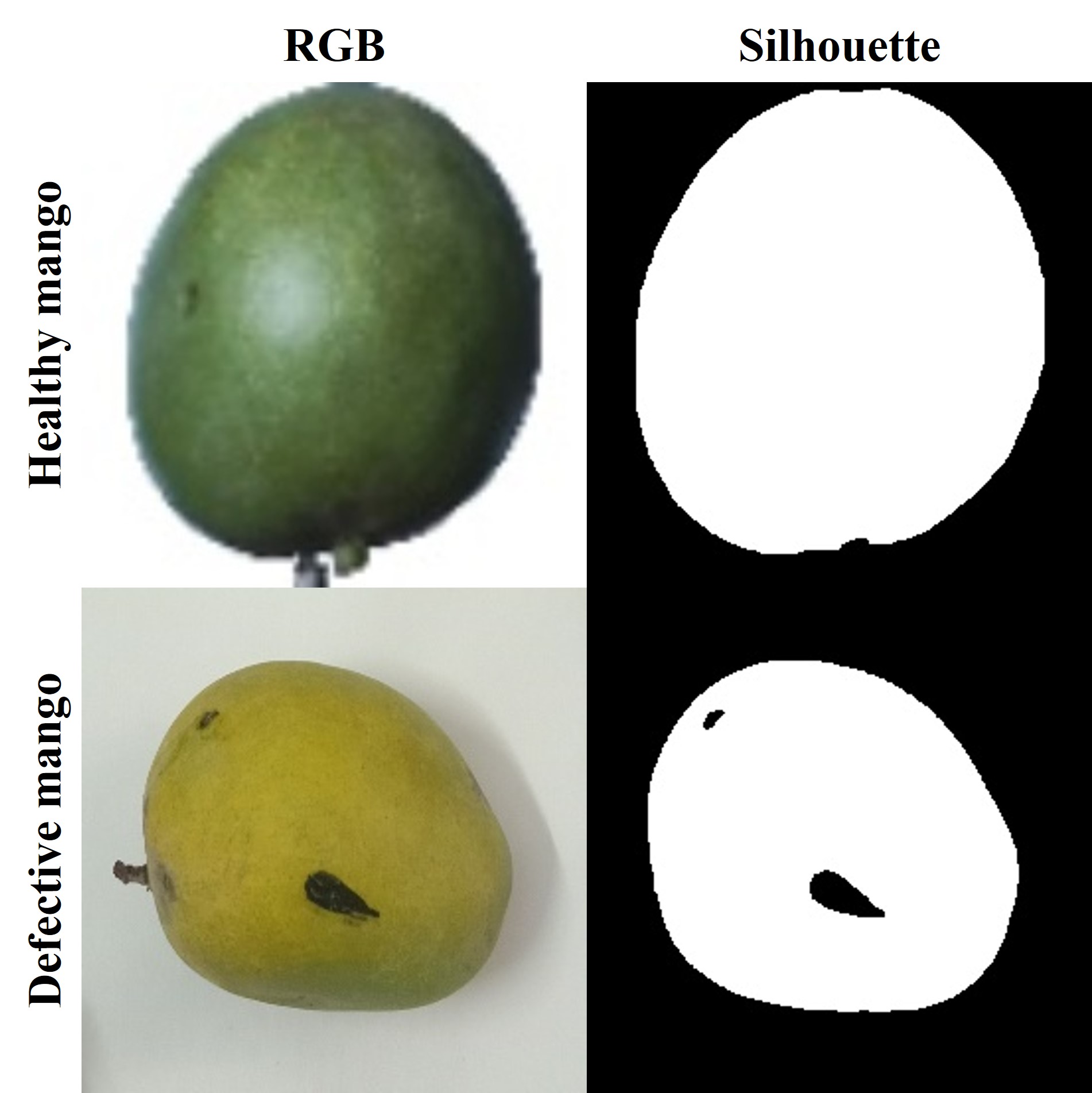}}
\caption{Apple silhouette.}
\label{fig_mango_sil}
\end{figure}

The acquired image dataset has been prepared for segmentation using the Segment Anything Model (SAM) tool~\cite{dikshit2023robochop}, allowing for the extraction of fruit silhouettes from the images. These silhouettes facilitate the identification of the object's shape. In our case, healthy fruits are binarized, and the complete silhouette of the fruit can be observed in white. Conversely, defective fruits are also binarized, but a difference can be observed when defects are present. These defects manifest in the white fruit silhouette with dark black pigmentations (see Fig.~\ref{fig_apple_sil} and Fig.~\ref{fig_mango_sil}).

After collecting the dataset consisting of 5k images per fruit, the SAM model was employed to generate segmented images. It was discovered that approximately 25\% of the images were segmented incorrectly. Consequently, a manual refinement of the dataset was carried out, resulting in the total number of images as indicated in Table~\ref{tab:datasetDistribution}.

\begin{table}[htbp]
\caption{Dataset distribution.}
\label{tab:datasetDistribution}
\begin{tabular}{@{}lllll@{}}
Fruits                 & Category                                     & Evaluation & \# Original Images & \# Refined Images \\ \hline
\multirow{6}{*}{Apple} & \multicolumn{1}{c}{\multirow{3}{*}{Healthy}} & Train      & 2000               & 1823              \\
                       & \multicolumn{1}{c}{}                         & Validation & 250                & 218               \\
                       & \multicolumn{1}{c}{}                         & Test       & 250                & 214               \\ \cline{2-5}
                       & \multirow{3}{*}{Defective}                   & Train      & 2000               & 1459               \\ 
                       &                                              & Validation & 250                & 170               \\
                       &                                              & Test       & 250                & 173               \\ \hline
\multirow{6}{*}{Mango} & \multicolumn{1}{c}{\multirow{3}{*}{Healthy}} & Train      & 2000               & 1624              \\
                       & \multicolumn{1}{c}{}                         & Validation & 250                & 203               \\
                       & \multicolumn{1}{c}{}                         & Test       & 250                & 221               \\ \cline{2-5}
                       & \multirow{3}{*}{Defective}                   & Train      & 2000               & 1478              \\
                       &                                              & Validation & 250                & 177               \\
                       &                                              & Test       & 250                & 202   \\ \hline           
\end{tabular}
\end{table}

\subsection{CNN Models Implemented}

For training (80\%), validation (10\%), and testing (10\%), the dataset is distributed as shown in Table~\ref{tab:datasetDistribution}. The images were organized into directories by fruit type (apple and mango), initially with 5k images each. However, after refining the dataset, the total number of apple images was reduced to 4,057 and the total number of mango images to 3,905.

The CNN models selected for this study are MobileNetV2 and VGG16, based on the state-of-the-art review.

MobileNetV2 has been specially designed for computer vision applications on devices with limited resources, such as mobile devices, which could be used by fruit inspectors, producers, or consumers in our case. This model stands out for its ability to significantly reduce the number of parameters and computational operations by using building blocks known as Inverted Residual Blocks, allowing it to maintain high performance in image classification, specifically in distinguishing between healthy and defective fruits. Additionally, MobileNetV2 employs a technique called Expansion Factor to control the number of channels in the intermediate layers of the network, facilitating the adaptation of the model to various tasks and computational resource requirements~\cite{xiang2019fruit}.

VGG16 is a deeply stacked model consisting of 16 layers of convolution and pooling, followed by three fully connected layers, and it is notable for its simplicity and relative depth. The model employs small-sized filters (3x3) in all convolutional layers, followed by max-pooling layers to reduce dimensionality. VGG16 has demonstrated good performance in image classification tasks on standard datasets, as is the case in our study~\cite{ukwuoma2022recent}.

\subsection{Multi-Input architecture}

The Multi-Input architecture using CNN models combines multiple streams of different types of input datasets, each processed by an independent CNN model, and as a final stage merges the learned features to perform a specific machine learning task~\cite{dua2021multi,choudhary2023multi,mesa2021multi}. The implemented Multi-Input architecture considers the utilization of two distinct input branches. In the first branch, an RGB dataset is employed, while in the second branch, a silhouette image dataset corresponding to each RGB image is utilized. Furthermore, independently trained models are deployed in each branch, namely MobileNetV2 and VGG16. Finally, the results obtained from each branch are unified through a feature fusion process, and a classification is achieved after a Multilayer Perceptron (MLP) layer. The following figure \ref{fig_multi} illustrates the methodology employed in this study.

\begin{figure}[htbp]
\centerline{\includegraphics[width=0.9\columnwidth]{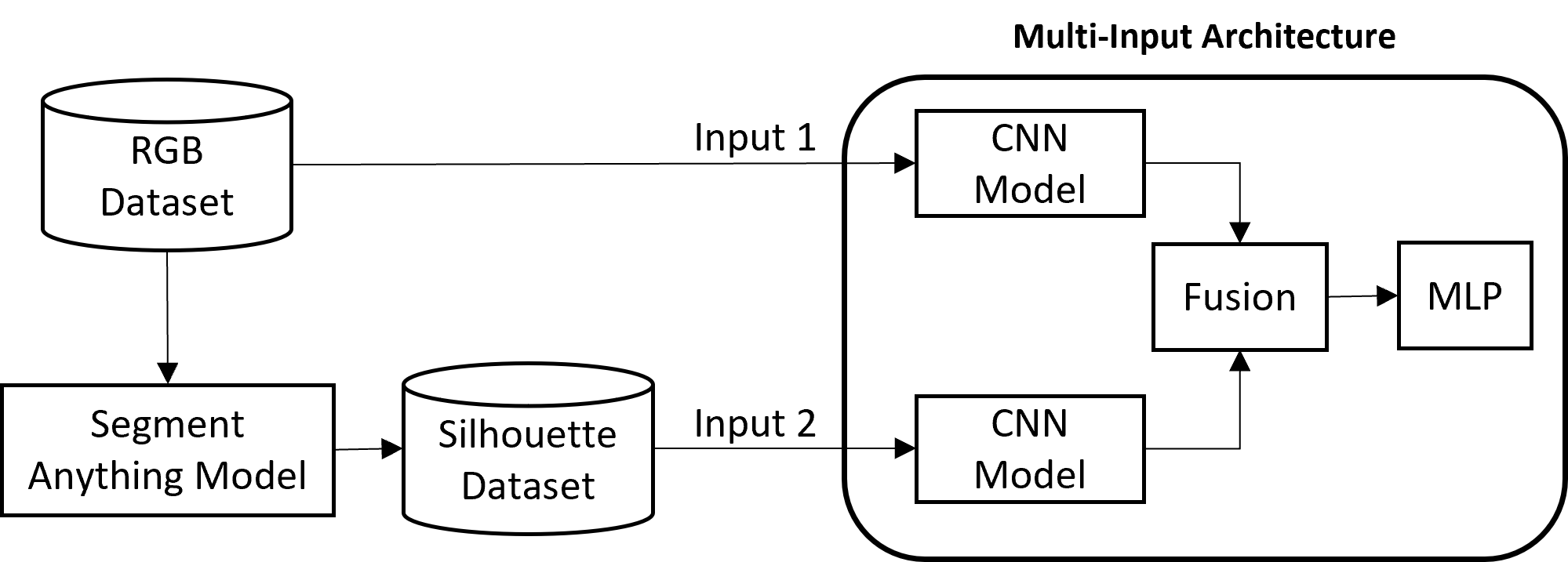}}
\caption{Multi-Input architecture used for healthy and defective fruits classification.}
\label{fig_multi}
\end{figure}

\section{Experimental Results}
\label{sec:experimental results}

This section outlines the outcomes derived from evaluating Multi-Input architectures in conjunction with CNN models, following the methodology previously elucidated. For each fruit type, the entire dataset, including their respective silhouettes, was utilized to assess the CNN models. To perform the comparison with the study~\cite{pacheco2023fruit}, the same fruits with their datasets and the types of defects they considered, namely rot, bruises, scabs, and black spots, are taken into account. It was noted that both models attained a 100\% accuracy rate in fruit classification. Nevertheless, upon reevaluation of the CNN models using refined datasets, equally high outcomes were achieved. In this instance, only the MobileNetV2 model achieved a 100\% accuracy rate.

The hyperparameters were established to train the architectures using input images sized 224x224. We employed the initial weights pre-trained with ImageNet for the CNN models. We opted for the Adam optimizer, with a low learning rate of 0.00001 and a batch size of 16. Additionally, a loss function designed for multiclass classification, categorical cross-entropy, was utilized, despite our classification being binary. Typically, training was configured to run up to epoch 60, yet it is noteworthy that due to achieving satisfactory results within the first 30 epochs, the number of epochs was reduced. As usual, the learned weights from the epoch exhibiting the best performance were selected as the best model.

\subsection{Results with Apples}

\begin{figure}[htbp]
\centerline{\includegraphics[width=0.9\columnwidth]{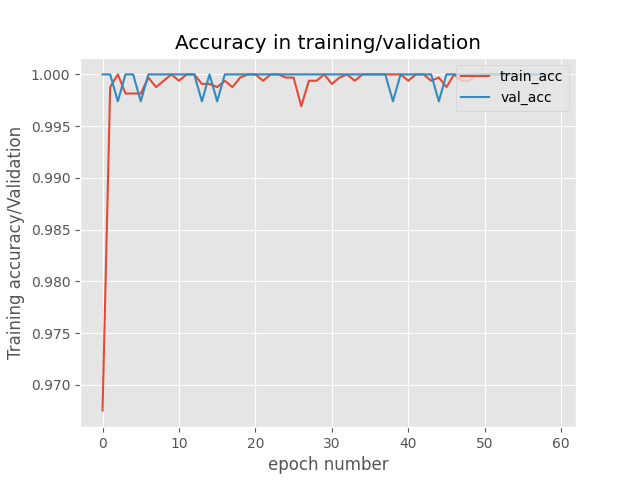}}
\caption{Multi-Input architecture:Training/Validation of the MobileNetV2 model with Apples.}
\label{fig:mobile_train_apple}
\end{figure}

\begin{figure}[htbp]
\centerline{\includegraphics[width=0.9\columnwidth]{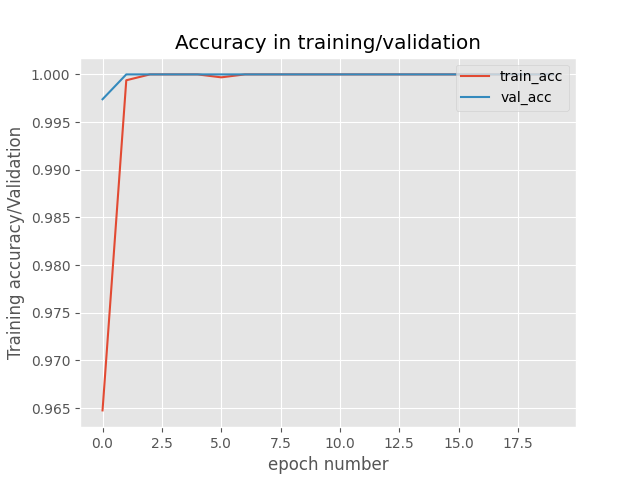}}
\caption{Multi-Input architecture:Training/Validation of the VGG16 model with Apples.}
\label{fig:vgg_train_apple}
\end{figure}

For the classification of healthy and defective apples, two CNN models were considered in training the Multi-Input architectures: MobileNetV2 and VGG16. Below, the training and validation plots for each model are presented (see Fig.~\ref{fig:mobile_train_apple} and Fig.~\ref{fig:vgg_train_apple}).

\begin{figure}[htbp]
\centerline{\includegraphics[width=0.9\columnwidth]{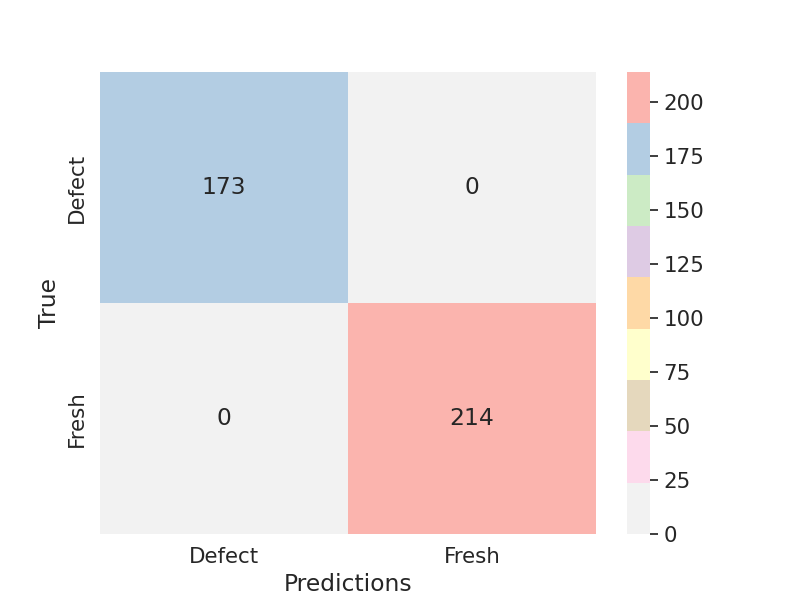}}
\caption{Multi-Input architecture: Confusion Matrix of the MobileNetV2 model with Apples.}
\label{fig:mobil_matrix_apple_Mobile}
\end{figure}

\begin{figure}[htbp]
\centerline{\includegraphics[width=0.9\columnwidth]{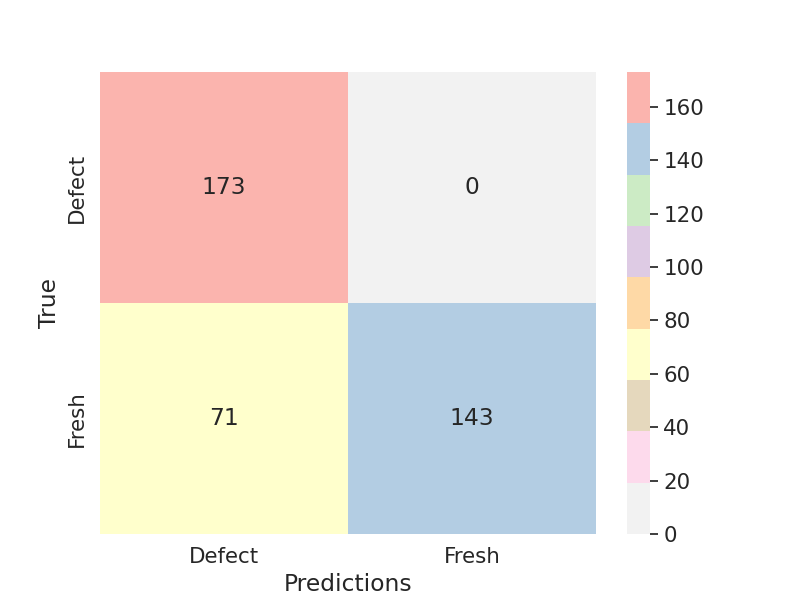}}
\caption{Multi-Input architecture: Confusion Matrix of the VGG16 model with Apples.}
\label{fig:mobil_matrix_apple_vgg}
\end{figure}

Upon reviewing the training and validation plots of the Multi-Input architectures, it is evident that they consistently achieve a high level of accuracy. This performance can be attributed to the contribution of silhouette images, as the segmentation process enables the Multi-Input architectures to accurately discern whether the fruit is healthy or defective. Furthermore, both Multi-Input architectures were subjected to testing, with the results depicted in Fig.~\ref{fig:mobil_matrix_apple_Mobile}, Fig.~\ref{fig:mobil_matrix_apple_vgg}, and Table~\ref{tab:results}.

During the testing phase of the Multi-Input architecture with the MobileNetV2 model, it became evident that all images of both healthy and defective apples were classified correctly, resulting in perfect accuracy. However, the Multi-Input architecture with the VGG16 model encountered challenges in accurately identifying images of healthy apples. This could be attributed to its design, which is optimized for classifying natural images containing rich details and textures. Consequently, it struggles when classifying silhouettes in binary images due to the loss of relevant information during conversion to a single channel and the lack of adaptation to the inherent simplicity of silhouettes.

\subsection{Results with Mangoes}

\begin{figure}[htbp]
\centerline{\includegraphics[width=0.9\columnwidth]{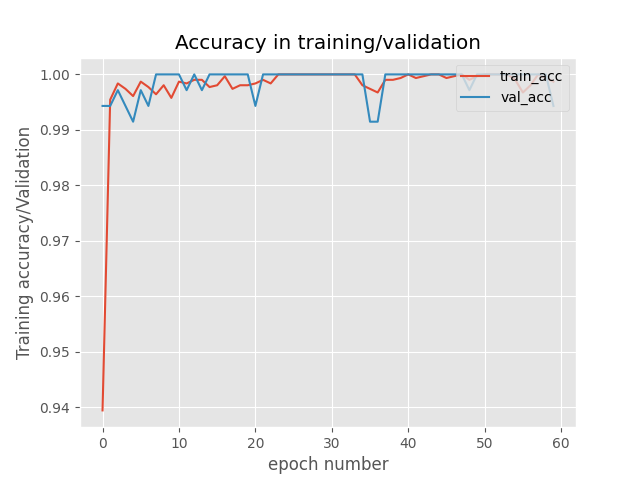}}
\caption{Multi-Input architecture: Training/Validation of the MobileNetV2 model with Mangoes.}
\label{fig:mobile_train_mango}
\end{figure}

\begin{figure}[htbp]
\centerline{\includegraphics[width=0.9\columnwidth]{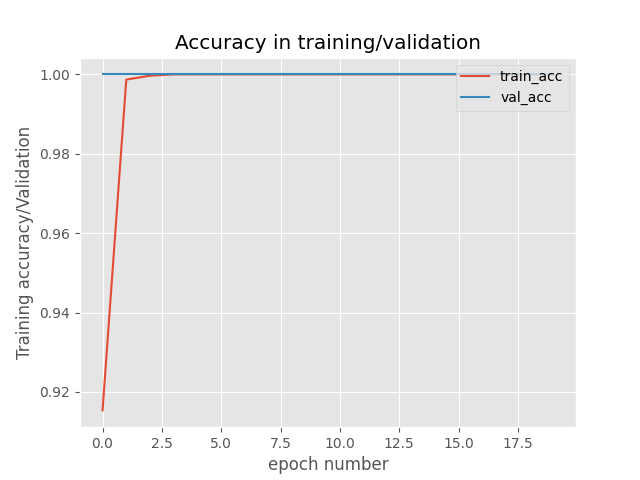}}
\caption{Multi-Input architecture: Training/Validation of the VGG16 model with Mangoes.}
\label{fig:vgg_train_mango}
\end{figure}

For the classification of healthy and defective mangoes, two CNN models were considered in training the Multi-Input architectures: MobileNetV2 and VGG16. Below, the training and validation plots for each model are presented (see Fig.~\ref{fig:mobile_train_mango} and Fig.~\ref{fig:vgg_train_mango}).

\begin{figure}[htbp]
\centerline{\includegraphics[width=0.9\columnwidth]{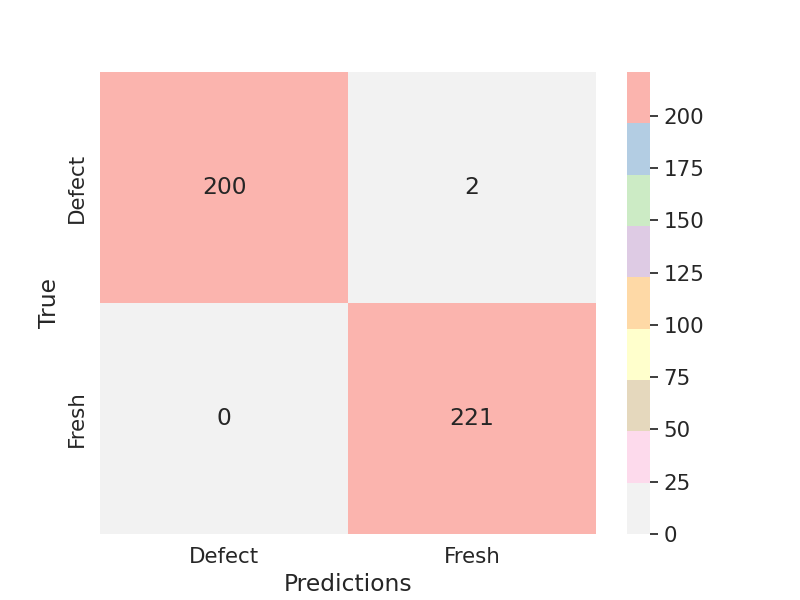}}
\caption{Multi-Input architecture: Confusion Matrix of the MobileNetV2 model with Mangoes.}
\label{fig:mobil_matrix_mango_Mobile}
\end{figure}

\begin{figure}[htbp]
\centerline{\includegraphics[width=0.9\columnwidth]{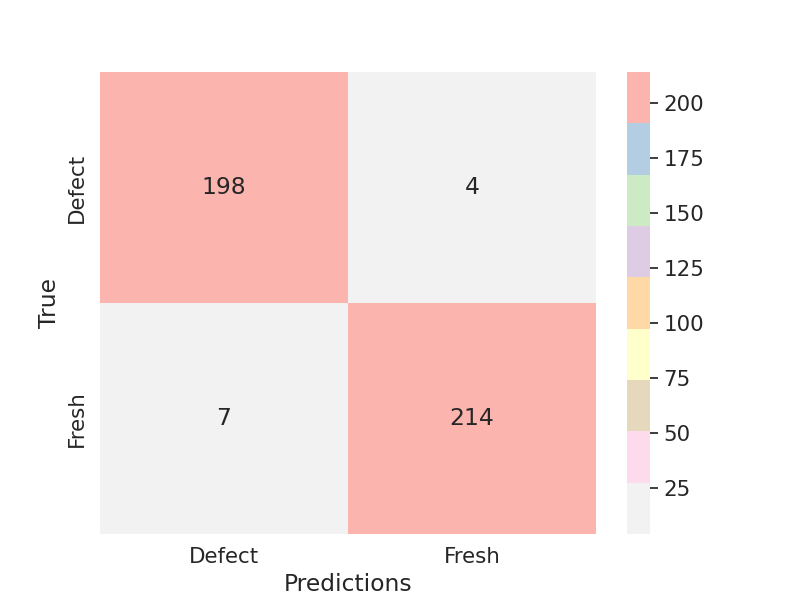}}
\caption{Multi-Input architecture: Confusion Matrix of the VGG16 model with Mangoes.}
\label{fig:mobil_matrix_mango_vgg}
\end{figure}

The training and validation plots of the Multi-Input architectures for the classification of healthy and defective mangoes exhibit a high level of accuracy, similar to what was observed with apples. This performance can also be attributed to the contribution of silhouette images, demonstrating the applicability of this technique across various fruit types. Furthermore, both Multi-Input architectures underwent testing, with the results depicted in Figures \ref{fig:mobil_matrix_mango_Mobile}, \ref{fig:mobil_matrix_mango_vgg}, and the Table~\ref{tab:results}.

Upon reviewing the confusion matrices from the testing phase of the Multi-Input architecture using the MobileNetV2 model, it was observed that nearly all images of defective mangoes were classified correctly, and all images of healthy mangoes in the test dataset were accurately classified, resulting in high accuracy. Conversely, the Multi-Input architecture with the VGG16 model correctly classified almost all images of both healthy and defective mangoes.

\begin{table*}[htbp]
\centering
\caption{Results Comparison.}
\label{tab:results}
\begin{tabular}{@{}llllll@{}}
\multicolumn{1}{c}{Fruits} & \multicolumn{1}{c}{Models} & \multicolumn{1}{c}{Accuracy} & Precision & Recall & F1-Score \\ \hline
\multirow{2}{*}{Apple}     & MobileNetV2 (Multi-Input)      & 1.0                          & 1.0       & 1.0    & 1.0      \\
                           & VGG16 (Multi-Input)            & 0.81653                      & 0.8545    & 0.8341 & 0.8154   \\ \hline
\multirow{2}{*}{Mango}     & MobileNetV2 (Multi-Input)      & 0.9953                       & 0.9955    & 0.9950 & 0.9953   \\
                           & VGG16 (Multi-Input)            & 0.9739                       & 0.9738    & 0.9743 & 0.9740  \\ \hline                    
\end{tabular}
\end{table*}

Upon reviewing the metrics obtained by the models in Table~\ref{tab:results}, it is evident that the best-performing model is MobileNetV2 (Multi-Input) for the classification of healthy and defective fruits. An accuracy of 100\% was achieved for apples and 99.53\% for mangoes. Comparing these metrics with those of the reviewed work~\cite{pacheco2023fruit}, which achieved 97.50\% accuracy for apples and 92.90\% for mangoes using only the MobileNetV2 model, it is clear that the MobileNetV2 (Multi-Input) model implemented in this study achieves higher efficiency than the models reviewed in the state of the art. It is worth mentioning that the same real dataset from the reviewed work was used. Furthermore, the MobileNetV2 (Multi-Input) model is not only suitable for silhouette images but also for any type of segmented binary images.

\section{Conclusions}
\label{sec:conclusions}

Robust models were successfully implemented and investigated, surpassing the performance reviewed in state-of-the-art, accurately classifying healthy or defective fruits using a Multi-Input architecture that utilizes both RGB and silhouette datasets in each branch. The inclusion of a binarized dataset (silhouette) significantly improved the performance of the Multi-Input architecture, enabling more precise classification and optimal performance. After comparing several models, it was determined that the best performer, for both apples and mangoes, is the MobileNetV2 (Multi-Input), achieving an accuracy of 100\% for apples and 99.53\% for mangoes. Furthermore, it is recommended for future work to implement this model and binary segmentation for the classification of other fruits, considering their healthy or defective state. 


\section*{Acknowledgment}
This work has been partially supported by the ESPOL-CIDIS-11-2022 project.

\bibliography{mybibfile}

\end{document}